\documentclass[letterpaper]{article}
\usepackage{aaai2027}
\usepackage[hyphens]{url}
\usepackage{graphicx}
\usepackage{natbib}
\usepackage{caption}
\usepackage{booktabs}

\title{MUCO: A Mani-Unit-Structured Manipulation Framework for Compositional Long-Horizon Robot Tasks}
\author{Xiangyi Wei, Hanzhe Yao, Jingwen Shang, Zhenhong Wu, Yang Li, Changbo Wang}
\affiliations{
    School of Computer Science and Technology, \\
    East China Normal University, Shanghai 200062, China \\
    \texttt{\{52285901014, 51295901117, 10235102513, 10235102482\}@stu.ecnu.edu.cn} \\
    \texttt{\{yli, cbwang\}@cs.ecnu.edu.cn} \\
    Corresponding authors: Yang Li, Changbo Wang
}

\begin{document}
\nocopyright

\maketitle

\begin{abstract}

Learning and executing long-horizon robot manipulation as complete task trajectories raises data-collection costs, limits capability reuse, and causes local failures to cascade into later task stages. We introduce manipulation units (MUs) as reusable operations, each defined by a local goal, an initiation condition, and observable success and failure criteria. MUs are learned and verified individually. During execution, consecutive MUs remain coupled through shared environment dynamics. We formulate MU-Structured Manipulation (MUSM) and realize it through MU-Centric Orchestration (MUCO). Given language and vision, MUCO constructs an MU sequence, matches each MU to compatible policy models in a frozen capability registry, and dispatches the model with the highest calibrated success rate. Visual verification checks MU outcomes against success and failure criteria and governs transitions, enabling retry or fallback to another model from the current state without re-executing verified MUs. Canonical keys enable checkpoint reuse for equivalent MUs across tasks without new demonstrations or updates. We evaluate MUCO on RLBench, RoboTwin 2.0, and five real-world tasks. MUCO achieves higher complete-task success and verified progress, stronger failure containment, and lower error propagation than baselines. Results demonstrate cross-task MU reuse and heterogeneous policy composition without joint training, shared backbones, or common action representations.
\end{abstract}

\section{Introduction}

A robot that can grasp a container, open its lid, and pour its contents should be able to recombine those capabilities when asked to prepare a different drink.
Under the dominant trajectory-centric paradigm, however, a new ordering or combination is still treated as a new end-to-end task, typically requiring another complete demonstration or another task-level policy.
The limitation is not simply that current policies are too small or their action horizons too short.
The deeper problem is that the trajectory is asked to play too many roles at once. It is the unit of data collection and policy learning, as well as the unit of execution, evaluation, and recovery.
For long-horizon manipulation, however, that unit is too coarse-grained.

Vision-language-action (VLA) models and related multimodal policies have substantially broadened the behaviors that robots can express by coupling perception, language, and continuous action generation \cite{brohan2023rt2,kim2024openvla,physicalintelligence2025pi05,driess2023palme,jiang2023vima,openx2024}. Yet greater policy capacity does not by itself delineate where one operation ends, whether its intended state change was achieved, or which existing controller should handle the next operation. These questions are of particular importance in laboratory procedures, household activities, and assembly tasks, where an early error changes the physical context faced by every later action \cite{mees2022calvin,jiang2023vima}. Long-horizon competence therefore requires not only a policy that can act, but also an organization of operation that makes intermediate progress reusable and inspectable.

The trajectory-centric formulation couples semantically distinct operations into one data collection, learning, and execution problem. At the data level, collecting demonstrations for a lengthy task requires complete successful rollouts, even when most constituent operations have already appeared in other tasks \cite{openx2024,khazatsky2024droid}. A change in task order or final objective can therefore require collecting another full trajectory.
At the capability level, the trajectory-level formulation does not expose the operations learned by a policy model as units that can be selected, verified, and recombined. Therefore, knowing how to grasp, pour, and insert does not by itself enable a robot to execute those operations in a new long-horizon ordering \cite{mees2022calvin,jiang2023vima}.
At execution time, an early failure changes the state encountered by all later operations. Without explicit Mani-Unit boundaries, it is difficult to check whether each operation succeeded or to retry only the failed operation \cite{hu2026matters,zhang2026visualverification}. This problem arises because learning and execution are organized around complete trajectories rather than reusable Mani-Units.

To address these problems, we introduce manipulation units (MUs) as reusable operations, each defined by a local goal, an initiation condition, and observable success and failure criteria. An MU specifies what state change should be achieved, while the selected policy model determines how to achieve it through continuous robot actions. Building on MUs, we formulate MU-Structured Manipulation (MUSM), which represents long-horizon tasks as ordered MU compositions. In MUSM, individually learned and verified MUs are  recombined into different task sequences.
During execution, the outcome of each MU determines the initial state of the next, so consecutive MUs remain coupled through shared environment dynamics.
Data are collected and policy models are trained separately for each MU. The same MU can then be reused across different task sequences and executed by different compatible policy models.|
Compared with language-guided planners that primarily organize execution through high-level skill selection or program generation \cite{ahn2022saycan,liang2023code,huang2022innermonologue}, MUSM does more than generate an execution sequence. MUSM represents long-horizon tasks as ordered compositions of reusable and verifiable MUs, and different tasks are formed by recombining the same MUs. Each MU retains its own data provenance and observable success and failure criteria. MUSM therefore changes both how task sequences are constructed and the granularity at which data are collected, policies are trained, outcomes are verified, and learned capabilities are reused across tasks.
We realize MUSM through \textbf{MU-Centric Orchestration (MUCO)}, which comprises four components: an MU constructor that converts task instructions and visual observations into ordered MU records; a matcher backed by a frozen capability registry that identifies compatible policy models and ranks them using calibrated success rates; model-specific adapters that translate observations and MU records into each policy model's native input; and a visual verifier that evaluates success and failure criteria and governs MU transitions and recovery. During execution, MUCO dispatches the highest-ranked compatible model for the active MU. After verified success, MUCO advances to the next MU. After failure, MUCO retries the current model or falls back to another compatible model from the current environment state without re-executing verified MUs.
Our experiments test whether this Mani-Unit-centric organization addresses the limitations of trajectory-centric learning described above. We first examine whether a learned Mani-Unit can be reused in a new task. We then test whether assigning different policy models to different Mani-Units improves task success. Finally, we evaluate whether Mani-Unit verification and retry prevent an early error from affecting the rest of a long-horizon task. Our contributions are:
\begin{itemize}
    \item We identify trajectory granularity as a shared bottleneck in data collection, capability reuse, and failure handling, introduce manipulation units (MUs) as reusable and verifiable operations, and formulate MUSM to represent long-horizon tasks as ordered compositions of MUs that can be recombined across task sequences.
    
    \item We introduce MUCO, a concrete realization of MUSM that integrates an MU constructor, a matcher backed by a frozen capability registry, model-specific adapters, and a visual verifier for transition control and failure recovery, without jointly retraining the underlying policy models.
    
    \item We establish an evaluation protocol that separates individual policy-model competence from the central claims of cross-task MU reuse without additional demonstrations, heterogeneous policy composition, and robustness to MU-level failures.
\end{itemize}

\section{Related Work}

\textbf{Vision-language-action models.}
VLA models and related multimodal policies learn language-conditioned visuomotor behavior from robot demonstrations and large-scale vision-language data \cite{brohan2023rt2,kim2024openvla,physicalintelligence2025pi05,driess2023palme,jiang2023vima,openx2024}. Scaling these models improves the breadth and generality of individual executors, but does not prescribe how a learned capability should be reused across task compositions or replaced after a local failure. MUSM is complementary to such scaling: a VLA can be registered once as an executor for every Mani-Unit it supports, reused across tasks without task-specific architectural changes, and replaced by a specialist only where calibrated evidence indicates that the specialist is more reliable. The contribution is therefore at the learning-and-deployment interface around policies, rather than in a new action decoder.

\textbf{Hierarchical VLA orchestration.}
Hierarchical VLA agents pair a high-level VLM with a low-level VLA and study planner choice, controller steerability, termination, observation representation, and memory \cite{hu2026matters}. This line establishes that orchestration choices matter even when a strong controller is available. MUSM is complementary but changes the object being orchestrated. A conventional hierarchy typically issues transient subgoals to one low-level controller; MUSM maintains a Mani-Unit as a persistent entity with its own data provenance, admissibility and outcome criteria, and a set of independently trained compatible executors. MUCO can consequently assign different executors to different occurrences of the same Mani-Unit and reuse the Mani-Unit record across task compositions.

\textbf{Language-guided task organization.}
Language models have been used to select skills according to affordances, synthesize executable programs, and replan from environmental feedback \cite{ahn2022saycan,liang2023code,huang2022innermonologue}. These approaches establish the value of semantic reasoning above robot control. MUSM focuses on the learning and deployment unit beneath that reasoning. Rather than treating a skill call as an atomic symbolic action whose implementation is hidden from the planner, MUSM retains each Mani-Unit as a stochastic feedback-control problem with Mani-Unit-level data, explicit outcome criteria, and multiple possible executors. The organizer sequences Mani-Units, but does not abstract away their continuous execution or its uncertainty.

\textbf{Task planning and hierarchical reinforcement learning.}
Classical task planners reason over symbolic action schemas, predicates, and precondition--effect structures \cite{kaelbling1998planning}; hierarchical reinforcement learning and options introduce temporal abstraction through initiation conditions, intra-option policies, and termination rules \cite{sutton1999between,dietterich2000hierarchical}. MUSM can be embedded in this formalism by treating each Mani-Unit--executor pair as an option, so our claim is not a new temporal-abstraction calculus. The distinction is the factorization MUSM makes explicit. A semantic Mani-Unit is separated from the policy that executes it; one Mani-Unit may map to multiple heterogeneous executors; a shared observation-based verifier judges all of them; and Mani-Unit data and capability estimates are reused across tasks. Options provide a language for temporally extended control, but do not by themselves prescribe this Mani-Unit-level organization of data, verification, calibration, and cross-task deployment.

\section{Problem Formulation}

We consider a language-conditioned manipulation task in a shared partially observed controlled process,
\begin{equation}
\mathcal{M}=(\mathcal{S},\mathcal{A},\mathcal{O},P,\rho_0),
\label{eq:shared-controlled-process}
\end{equation}
where $s_t\in\mathcal{S}$ is the unobserved environment state, $o_t\in\mathcal{O}$ is the robot observation, $a_t\in\mathcal{A}$ is a continuous robot action, $P(s_{t+1}\mid s_t,a_t)$ is the shared transition dynamics, and $\rho_0$ is the initial-state distribution. The robot receives a language instruction $\ell$ and must satisfy its task-level completion condition. MUSM organizes trained executors within this shared environment without factorizing $P$ or assuming that the state produced by one MU is irrelevant to the next.

Within this shared controlled process, MUSM associates the task with an ordered sequence of MUs $\mathbf{u}=(u_1,\ldots,u_K)$. An MU is defined as
\begin{equation}
u_i=(g_i,\mathcal{S}_i,\mathcal{A}_i,\mathcal{I}_i,v_i,\Pi_i),
\end{equation}
where $g_i$ is a local goal, $\mathcal{S}_i\subseteq\mathcal{S}$ and $\mathcal{A}_i\subseteq\mathcal{A}$ specify the state and action components relevant to that goal, $\mathcal{I}_i$ is an initiation condition, $v_i$ returns $\{\textsc{success},\textsc{failure},\textsc{continue}\}$, and $\Pi_i$ is a set of compatible stochastic execution policies. Policies in $\Pi_i$ may differ in architecture, training data, or control precision, provided that they can pursue $g_i$ through an interface compatible with the MU. The evolution of $u_i$ remains governed by the global dynamics $P$ over the full state, rather than by an independent MU-specific kernel. Each fixed pair $(u_i,\pi_k)$ is representable as an option; further discussion is provided in the supplementary material.

\paragraph{Boundary-Level Organization.}

Let $i_m\in\{1,\ldots,K\}$ denote the position of the active MU at the $m$th decision boundary $\tau_m$, and let $h_{\tau_m}$ contain all earlier MU--policy attempts, verification outcomes, and boundary observations. Separating $i_m$ from $m$ is important: a failed attempt creates a new decision boundary without advancing the ordered MU plan. Let $\Pi_{i_m}^{(m)}\subseteq\Pi_{i_m}$ be the policies still eligible after the preceding attempts. At each boundary, an organizer $\omega$ selects an executor for the \emph{active} MU using the current observation, task instruction, and history:
\begin{equation}
k_m \sim \omega(\cdot \mid u_{i_m},o_{\tau_m},\ell,h_{\tau_m}),
\qquad \pi_{k_m}\in\Pi_{i_m}^{(m)}.
\end{equation}
The active MU must satisfy the explicit admissibility condition
\begin{equation}
\mathcal{I}_{i_m}(o_{\tau_m},\ell,h_{\tau_m})=1.
\end{equation}
The selected executor generates stochastic continuous controls while the MU remains active,
\begin{equation}
\begin{array}{rcl}
a_t &\sim& \pi_{k_m}(\cdot\mid o_t,u_{i_m}),\\
&& \tau_m\leq t<\tau_{m+1}.
\end{array}
\end{equation}
where $\tau_{m+1}$ is the first time at which $v_{i_m}$ returns \textsc{success} or \textsc{failure}. On success, $i_{m+1}=i_m+1$. On failure, the organizer retains $i_{m+1}=i_m$ and updates $\Pi_{i_m}^{(m)}$ to implement retry or policy fallback from the current global state; it aborts when no eligible policy remains. This formulation preserves the MU order supplied by the instruction rather than assuming unconstrained task replanning. It introduces explicit verification and executor-reassignment points into a global, state-dependent continuous-control problem.

The task succeeds only when every MU in the instruction reaches a
verified \textsc{success} boundary:
\begin{equation}
\mathrm{Succ}(\ell)=
\mathbf{1}\!\left[
\bigwedge_{i=1}^{K}
\exists m:\ i_m=i\ \land\
v_i(o_{\tau_{m+1}})=\textsc{success}
\right].
\label{eq:task-success}
\end{equation}
Let $\mathcal{D}_{\mathrm{task}}$ denote the distribution over
language-conditioned task instances. The organizer's task-level objective
is to maximize long-horizon task success:
\begin{equation}
\begin{array}{rcl}
\omega^{*} &\in& \displaystyle\arg\max_{\omega}J(\omega),\\
J(\omega) &=& \mathrm{E}_{\omega}\!\left[\mathrm{Succ}(\ell)\right]\\
&=& \Pr_{\omega}\!\left[\mathrm{Succ}(\ell)=1\right].
\end{array}
\label{eq:task-objective}
\end{equation}
The expectation is over $\ell\sim\mathcal{D}_{\mathrm{task}}$,
$s_0\sim\rho_0$, the shared dynamics $P$, and stochastic executor
rollouts. MU-level retry and executor fallback are mechanisms for
improving this task-level objective rather than separate objectives.

\section{Method}

MUCO realizes MUSM through four components. A Mani-Unit constructor converts task instructions into shared Mani-Unit records, an offline registry links Mani-Units to compatible executors and calibrated success rates, model-specific adapters translate shared inputs into each executor's native inputs, and a visual verifier controls Mani-Unit transitions and recovery. Executors act at their native control frequencies, while MUCO intervenes only at Mani-Unit boundaries and every $\Delta$ seconds for verification. This design coordinates heterogeneous policies without joint training, a shared backbone, or a common action representation.

\subsection{Instruction-Grounded Mani-Unit Records}

At time $t$, MUCO receives the three-view RGB observation
$\mathbf{I}_t=(I_t^{\mathrm{left}},I_t^{\mathrm{front}},
I_t^{\mathrm{right}})$, consisting of images from the left-wrist, front,
and right-wrist cameras,
respectively. This observation forms the visual component of $o_t$. Given
a task instruction $\ell$ and the initial three-view observation
$\mathbf{I}_0$, the Mani-Unit constructor produces an ordered sequence of
implementation records:
\begin{equation}
\begin{array}{rcl}
F_{\mathrm{unit}}(\ell,\mathbf{I}_0)
&=&(\xi_1,\ldots,\xi_K),\\
\xi_i&=&(z_i,\ell_i,c_i^{+},c_i^{-},T_i).
\end{array}
\end{equation}
Here, $z_i\in\mathcal{Z}$ identifies a canonical Mani-Unit, $\ell_i$ describes the required local state change, $c_i^{+}$ and $c_i^{-}$ specify observable success and failure conditions, and $T_i$ sets the maximum duration. The three-view observation $\mathbf{I}_0$ grounds each record in the current scene, while $z_i$ links equivalent Mani-Unit descriptions to the same capability entry.

We implement the semantic part of $F_{\mathrm{unit}}$ using a frozen
multimodal language model with a fixed output schema. Given
$(\ell,\mathbf{I}_0)$ and $\mathcal{Z}$, the model returns the canonical key,
operation description, and visual success and failure criteria for each
Mani-Unit; the constructor appends the task-configured $T_i$. Restricting $z_i$
to $\mathcal{Z}$ allows the registry, dispatcher, and verifier to use the same
Mani-Unit record consistently.

In the MUSM formulation, $\ell_i$ specifies $g_i$, while $(c_i^{+},c_i^{-},T_i)$ defines $v_i$. The remaining elements are instantiated by the evaluator, registry, and executor adapters. At decision boundary $\tau_m$, the evaluator computes
\begin{equation}
\begin{array}{rcl}
\iota_m
&=&F_{\mathrm{init}}(\mathbf{I}_{\tau_m},\ell,\xi_{i_m},h_{\tau_m})\\
&=&\mathcal{I}_{i_m}(o_{\tau_m},\ell,h_{\tau_m})
\in\{0,1\}.
\end{array}
\end{equation}
Using the same frozen model with a binary output schema, $F_{\mathrm{init}}$ checks whether the active Mani-Unit can begin. MUCO dispatches the Mani-Unit only when $\iota_m=1$. After a failed attempt, it evaluates the latest observation again so that recovery starts from the observed state after failure.

\subsection{Offline Mani-Unit--Executor Capability Registry}

Let $\mathcal{E}=\{e_k\}_{k=1}^{M}$ be a library of independently trained
executors, where $e_k$ implements stochastic policy $\pi_k$. Each registry
entry stores an identifier, a capability and interface descriptor $b_k$, a
launcher, a model-specific adapter $\psi_k$, and a frozen checkpoint. The
descriptor $b_k$ contains canonical Mani-Unit keys, language aliases, and
required observation and action fields. This common description covers
policies that use only sensor observations as well as VLAs that also receive
$\ell_i$.

Executors that support the same Mani-Unit may differ in reliability. MUCO therefore calibrates each compatible pairing of a Mani-Unit and an executor before evaluation. For pair $(z,k)$, the registry records $N_{zk}$ trials, $S_{zk}$ successes, and the empirical success rate
\begin{equation}
\widehat q_{zk}=\frac{S_{zk}}{N_{zk}}.
\end{equation}
Simulation calibration uses privileged environment success checks. For physical calibration, human reviewers correct the model's visual judgments when needed. These outcomes are used only to build the offline registry and are never available to the online organizer. The registry remains frozen throughout evaluation.

For a Mani-Unit record $\xi_i$, a matcher compares the canonical key, local
description, and interface requirements with each descriptor $b_k$. The
compatible executor set is
\begin{equation}
\mathcal{E}_i=
\{e_k\in\mathcal{E}:F_{\mathrm{match}}(z_i,\ell_i,b_k)=1\}.
\end{equation}
The policies associated with $\mathcal{E}_i$ instantiate the compatible set
$\Pi_i$ in the MUSM formulation. We implement $F_{\mathrm{match}}$ with the
language branch of the same frozen model and a schema that requires a binary
output. The matcher uses the Mani-Unit description rather than exact string
matching. Compatibility determines which executors can be used, while
$\widehat q_{zk}$ determines their ranking.

\subsection{Capability-Aware Dispatch and Adaptation}

At decision boundary $\tau_m$, let
$B_{i_m}^{(m)}\subseteq\mathcal{E}_{i_m}$ contain the executors whose retry
budgets for the active Mani-Unit have been exhausted. The policies associated
with $\mathcal{E}_{i_m}\setminus B_{i_m}^{(m)}$ instantiate the remaining
eligible set $\Pi_{i_m}^{(m)}$ in the MUSM formulation. When this set is
nonempty, MUCO selects the eligible executor with the highest calibrated
success rate:
\begin{equation}
k_m^{*}=\arg\max_{k:e_k\in
\mathcal{E}_{i_m}\setminus B_{i_m}^{(m)}}
\widehat q_{z_{i_m}k}.
\end{equation}
The dispatcher retrieves the selected entry's launcher, adapter, and
checkpoint. It starts the executor, and the adapter converts the current
observation and Mani-Unit record into the executor's native input:
\begin{equation}
a_t\sim\pi_{k_m^{*}}\!\left(
\cdot\mid\psi_{k_m^{*}}(o_t,\xi_{i_m})
\right).
\end{equation}
Selection is therefore based on measured reliability for the active Mani-Unit. The adapter allows policies with different observation formats, action chunk sizes, and control rates to retain their native interfaces without parameter sharing or retraining.

The selected executor remains fixed during an attempt, so organizer decisions do not interrupt fast feedback control. After a verified failure, MUCO records the executor, outcome, and boundary observation before retrying or choosing the eligible executor with the next highest score. This fallback handles individual outcomes that differ from the calibrated rates and requires no learned meta policy.

\begin{table}[t]
\centering
\small
\caption{Mani-Unit transitions controlled by verification. Recovery never replays an already verified Mani-Unit.}
\label{tab:unit-transitions}
\begin{tabular}{@{}p{0.60in}p{0.86in}p{1.35in}@{}}
\toprule
Signal & Condition & Transition \\
\midrule
\textsc{continue} & any & keep current executor \\
\textsc{success} & $i<K$ & record boundary; activate $i+1$ \\
\textsc{success} & $i=K$ & declare task success \\
\textsc{failure} & retry available & retry current executor \\
\textsc{failure} & fallback available & dispatch next executor \\
\textsc{failure} & otherwise & declare task failure \\
\bottomrule
\end{tabular}
\end{table}

\subsection{Verification-Gated Transition and Local Recovery}

During attempt $m$, the selected policy runs continuously from its native observation stream. Every $\Delta$ seconds, MUCO samples the current three-view RGB observation $\mathbf{I}_t$ and evaluates the active Mani-Unit
\begin{equation}
\begin{array}{rcl}
y_{i_m,t}&=&V(\mathbf{I}_t,c_{i_m}^{+},c_{i_m}^{-},
t-\tau_m,T_{i_m})\\
&\in&\{\textsc{continue},\textsc{success},\textsc{failure}\}.
\end{array}
\end{equation}
The verifier returns \textsc{success} when $c_{i_m}^{+}$ is visually
satisfied, \textsc{failure} when $c_{i_m}^{-}$ is satisfied or the elapsed
time $t-\tau_m$ exceeds $T_{i_m}$, and \textsc{continue} otherwise. It uses
the same three-view RGB interface in simulation and on the physical robot.
Simulator success signals are used only for offline calibration and final
evaluation, never for online transitions.

At every decision boundary $\tau_m$, the active Mani-Unit starts from the
physical boundary state $s_{\tau_m}$ produced by the complete preceding
execution history, including verified Mani-Units and any failed attempts.
Although this state is only partially observed, retries and executor
fallback both continue from the current global state rather than from a
reset or local Mani-Unit initial state. Mani-Unit execution and recovery
therefore remain governed by the shared transition dynamics $P$ of the
controlled process in Eq.~(\ref{eq:shared-controlled-process}), preserving
physical coupling across Mani-Unit boundaries.

Table~\ref{tab:unit-transitions} summarizes the resulting state machine. On \textsc{success}, MUCO records the verified boundary in $h_{\tau_{m+1}}$, advances to $i_{m+1}=i_m+1$, and checks whether the next Mani-Unit can begin. On \textsc{failure}, it records the boundary, retains $i_{m+1}=i_m$, and checks the active Mani-Unit again using the latest observation and history. If the Mani-Unit can still begin and the executor has retries remaining, MUCO restarts it from the current state. Otherwise, the executor is added to $B_{i_m}^{(m+1)}$ and the dispatcher selects the best remaining alternative. The task fails when the active Mani-Unit can no longer begin or no eligible executor remains.

Recovery applies only to the active Mani-Unit, so previously verified
Mani-Units are not replayed. Every recovery attempt must nevertheless pass
the initiation check again. The realized outcomes
$y_{i_m,\tau_{m+1}}$ instantiate
$v_{i_m}(o_{\tau_{m+1}})$ in the task-success event of
Eq.~(\ref{eq:task-success}).

\section{Experiments}

\begin{table*}[t]
\centering
\caption{Simulation task success rate (\%). (a) RLBench prior values are reported by LaST0~\cite{last0} and DeepVision-VLA~\cite{deepvisionvla}. (b) RoboTwin 2.0 results are placeholders. Higher is better.}
\label{tab:simulation-results}
\resizebox{\textwidth}{!}{%
\begin{tabular}{lccccccccccc}
\toprule
\multicolumn{12}{l}{\textit{(a) RLBench}} \\
Method & Box & Laptop & Toilet & Sweep & Fridge & Phone & Umbrella & Frame & Wine & Water & Mean \\
\midrule
OpenVLA & 60 & 35 & 75 & 55 & 85 & 20 & 30 & 15 & 20 & 5 & 40 \\
SpatialVLA & 80 & 70 & 85 & 20 & 80 & 15 & 25 & 40 & 15 & 30 & 46 \\
CogACT & 90 & 80 & 95 & 50 & 85 & 50 & 55 & 45 & 30 & 25 & 61 \\
CoT-VLA & 95 & 75 & 100 & 80 & 65 & 50 & 40 & 50 & 55 & 50 & 66 \\
$\pi_{0.5}$ & 90 & 95 & 85 & 75 & 100 & 5 & 10 & 80 & 75 & 35 & 65 \\
HybridVLA & 85 & 95 & 100 & 90 & 100 & 50 & 50 & 70 & 50 & 50 & 74 \\
QwenVLA-OFT & 95 & 95 & 100 & 15 & 95 & 65 & 30 & 70 & 65 & 50 & 69 \\
LaST0 & 95 & 95 & 100 & 80 & 85 & 75 & 75 & 70 & 85 & 60 & 82 \\
DeepVision-VLA & 100 & 95 & 100 & 95 & 95 & 75 & 65 & 70 & 85 & 50 & 83 \\
\midrule
\textbf{MUCO (ours)} & -- & -- & -- & -- & -- & -- & -- & -- & -- & -- & -- \\
\midrule
\multicolumn{12}{l}{\textit{(b) RoboTwin 2.0}} \\
Method & RT-1 & RT-2 & RT-3 & RT-4 & Mean & \multicolumn{6}{c}{} \\
X-VLA & -- & -- & -- & -- & -- & \multicolumn{6}{c}{} \\
Executor $e_1$ & -- & -- & -- & -- & -- & \multicolumn{6}{c}{} \\
Executor $e_2$ & -- & -- & -- & -- & -- & \multicolumn{6}{c}{} \\
Executor $e_3$ & -- & -- & -- & -- & -- & \multicolumn{6}{c}{} \\
\textbf{MUCO (ours)} & -- & -- & -- & -- & -- & \multicolumn{6}{c}{} \\
\bottomrule
\end{tabular}%
}
\end{table*}

\begin{table*}[t]
\centering
\small
\caption{H1 results. SR is complete-task success, Prog is normalized verified progress, EPC is error propagation, and FCR is failure containment. All entries are placeholders.}
\label{tab:h1-results}
\begin{tabular}{lcccccccc}
\toprule
Method & Short SR & Task-A SR & Task-B SR & General SR & Mean SR & Prog & EPC & FCR \\
\midrule
Flat-Same & -- & -- & -- & -- & -- & -- & -- & -- \\
Trajectory-E2E & -- & -- & -- & -- & -- & -- & -- & -- \\
\textbf{MUCO-Same} & -- & -- & -- & -- & -- & -- & -- & -- \\
\bottomrule
\end{tabular}
\end{table*}

\subsection{Evaluation Questions and Protocol}

Our experiments test three hypotheses that follow directly from the Mani-Unit--executor factorization. \textbf{H1 (Mani-Unit organization):} with low-level policy capacity controlled, Mani-Unit decomposition, explicit verification, and local retry or fallback improve long-horizon task completion. \textbf{H2 (Mani-Unit reuse):} an executor trained for a Mani-Unit can be reused, without additional Mani-Unit demonstrations or checkpoint updates, when that Mani-Unit appears in a new task composition. \textbf{H3 (heterogeneous composition):} selecting a different specialist for each Mani-Unit outperforms forcing one policy model to execute every Mani-Unit. We additionally test whether these conclusions are robust to the multimodal agent used by MUCO.

\paragraph{Terminology.}
We reserve \emph{Mani-Unit} for the reusable manipulation unit defined in Section~3 and \emph{executor} for a concrete policy checkpoint that can realize it. We use \emph{Mani-Unit-level policy} only when emphasizing training granularity. This avoids overloading \emph{skill}, \emph{option}, or \emph{action}, which refer to different abstractions in prior work.

\paragraph{Metrics and statistical protocol.}
Task success requires every Mani-Unit to reach a verified \textsc{success} boundary; normalized progress is the fraction of Mani-Units verified before termination. We also report per-Mani-Unit success, low-level actions, error propagation $\Pr(F_{i+1}\mid F_i, i+1\ \mathrm{attempted})$, and failure containment, the fraction of failed first attempts recovered without replaying a verified prefix. Unless an official protocol differs, each configuration uses the same 20 initial conditions; success rates include 95\% Wilson intervals and paired differences.

\subsection{Simulation Experiments}

\subsubsection{RLBench: Standard Task Competence}

\paragraph{Role of the benchmark.}
RLBench~\cite{james2020rlbench} provides a standardized comparison with strong VLA policies, but its ten tasks are predominantly short horizon. We therefore use it to establish that Mani-Unit-structured execution preserves standard manipulation competence, rather than as the primary evidence for long-horizon composition. We evaluate Close box, Close laptop lid, Toilet seat down, Sweep to dustpan, Close fridge, Phone on base, Take umbrella out, Frame off hanger, Wine at rack, and Water plants. Following LaST0 and DeepVision-VLA, the robot is a Franka Panda in CoppeliaSim with a single front-view RGB observation. The reported protocol uses 100 waypoint-generated demonstrations per task, 20 rollout trials per task, and three random seeds.

\paragraph{Comparison methods.}
The comparison spans OpenVLA~\cite{kim2024openvla}, SpatialVLA~\cite{qu2025spatialvla}, CogACT~\cite{li2024cogact}, CoT-VLA~\cite{zhao2025cotvla}, $\pi_{0.5}$~\cite{physicalintelligence2025pi05}, HybridVLA~\cite{liu2025hybridvla}, LaST0~\cite{last0}, and DeepVision-VLA~\cite{deepvisionvla}. Prior RLBench values in Table~\ref{tab:simulation-results}(a) are transcribed from LaST0 and DeepVision-VLA rather than re-evaluated in our codebase; the exact $\pi_{0.5}$ update count is not disclosed by those reports.

\subsubsection{RoboTwin: Bimanual and Compositional Simulation}

RoboTwin 2.0~\cite{chen2025robotwin} complements RLBench with bimanual manipulation and strong domain randomization. We use a fixed four-task subset for Mani-Unit composition and heterogeneous dispatch, with 20 matched rollouts per method. Exact task names, decompositions, and demonstration budgets will be stated in the final protocol. Table~\ref{tab:simulation-results}(b) compares the executor library, MUCO, and X-VLA~\cite{zheng2026xvla}, which provides an official RoboTwin 2.0 checkpoint.

\subsection{Real-World Experiments}

\subsubsection{Platform, Mani-Unit Library, and Tasks}

All real-world configurations use the same bimanual robot platform, camera placement, object set, and low-level control interface. These factors are held fixed across methods so that comparisons change only the data and execution organization stated below. Each executor is trained independently on demonstrations of the Mani-Units it supports; no executor is jointly trained with the MUCO organizer.

The canonical Mani-Unit vocabulary contains six entries: $A$, \textsc{Transfer}, grasps an object and moves it to the work area; $B$, \textsc{Pour}, transfers contents between containers; $C$, \textsc{Titrate}, dispenses reagent with a burette; $D$, \textsc{Press}, activates an apparatus; $E$, \textsc{Carry}, moves an object to a target table; and $F$, \textsc{Insert}, aligns and inserts an object into a receptacle or fixture. \textsc{Insert} provides a general contact-rich operation beyond liquid handling. The library contains independently trained executors $e_1,e_2,e_3$ and, where interfaces are compatible, X-VLA; the final paper will report their model families, supported keys, and demonstration counts per Mani-Unit in the implementation details rather than a separate table.

Before formal evaluation, every compatible Mani-Unit--executor pair is tested in $N_{\mathrm{cal}}=20$ independent calibration trials. Calibration initial states are disjoint from both training demonstrations and formal test configurations. The resulting $\widehat q_{zk}$ values and compatibility masks are frozen before any long-horizon trial.

We evaluate four compositions: Task-Short, $A\rightarrow C$; Task-A, $A\rightarrow B\rightarrow C\rightarrow E\rightarrow D$; Task-B, $A\rightarrow C\rightarrow B\rightarrow D\rightarrow E$; and Task-General, $A\rightarrow F\rightarrow D\rightarrow E$. No complete trajectory from Task-A, Task-B, or Task-General is used to train an executor for a Mani-Unit.

\subsubsection{H1: Mani-Unit Organization versus Trajectory Coupling}

This experiment isolates Mani-Unit organization from low-level policy capacity. A single comparison cannot simultaneously use an identical checkpoint and different training granularity, so we report two controls. \textsc{Flat-Same} uses the same executor checkpoint as MUCO; this checkpoint is trained at Mani-Unit granularity, but Flat-Same disables boundary verification, retry, and fallback. This is the organization-only ablation. \textsc{Trajectory-E2E} trains one task-level policy from complete trajectories using a matched interaction-frame budget; this compares the full trajectory-centric and Mani-Unit-centric formulations. \textsc{MUCO-Same} forces every Mani-Unit to use the same executor, preserving Mani-Unit construction, visual verification, and retry with $R=3$ while removing heterogeneous selection.

\subsubsection{Independent Mani-Unit Capability}

We next execute each Mani-Unit from a standardized admissible initial state, without preceding or subsequent Mani-Units. Each compatible $(z,k)$ pair is evaluated for 20 trials. The matrix in Table~\ref{tab:unit-capability} serves two roles: it measures whether individual executors are reliable on their intended Mani-Units, and it provides the frozen capability evidence used to interpret heterogeneous selection. Non-compatible pairs are marked N/A rather than counted as failures.

\begin{table}[t]
\centering
\small
\setlength{\tabcolsep}{3pt}
\caption{Independent Mani-Unit success rate (\%).}
\label{tab:unit-capability}
\begin{tabular}{lcccccc}
\toprule
Executor & A & B & C & D & E & F \\
\midrule
$e_1$ & -- & -- & -- & -- & -- & -- \\
$e_2$ & -- & -- & -- & -- & -- & -- \\
$e_3$ & -- & -- & -- & -- & -- & -- \\
X-VLA & -- & -- & -- & -- & -- & -- \\
\bottomrule
\end{tabular}
\end{table}

For keys $(z_1,\ldots,z_K)$, the product $\widehat U_{\mathrm{ind}}=\prod_i\max_k\widehat q_{z_i k}$ is an \emph{independence estimate}, not an upper bound, because shared-state dependencies correlate Mani-Unit outcomes.

\subsubsection{H2--H3: Reuse and Heterogeneous Composition}

To test H3, all configurations retain the same constructor, verifier, retry budget, and Mani-Unit data. MUCO-Het selects $\arg\max_k\widehat q_{z_i k}$, whereas each homogeneous ablation forces one executor across all compatible Mani-Units. Table~\ref{tab:h2h3-results}(a) therefore isolates heterogeneous dispatch from Mani-Unit organization.

\begin{table}[t]
\centering
\scriptsize
\setlength{\tabcolsep}{2.5pt}
\caption{Heterogeneous composition and cross-task reuse. (a) H3 task success rate under different executor rules. (b) H2 reuses the same checkpoint with zero new Mani-Unit demonstrations. All result entries are placeholders.}
\label{tab:h2h3-results}
\begin{tabular}{lccccc}
\toprule
\multicolumn{6}{l}{\textit{(a) H3: heterogeneous versus homogeneous execution}} \\
Method & Rule & A SR & B SR & Mean & V/R \\
\midrule
MUCO-$e_1$ & always $e_1$ & -- & -- & -- & Yes \\
MUCO-$e_2$ & always $e_2$ & -- & -- & -- & Yes \\
MUCO-$e_3$ & always $e_3$ & -- & -- & -- & Yes \\
\textbf{MUCO-Het} & $\arg\max_k\widehat q_{z_i k}$ & -- & -- & -- & Yes \\
\midrule
\multicolumn{6}{l}{\textit{(b) H2: cross-task Mani-Unit reuse}} \\
Mani-Unit & Exec. & A SR & B SR & $|\Delta|$ & New \\
\midrule
Transfer (A) & -- & -- & -- & -- & 0 \\
Titrate (C) & -- & -- & -- & -- & 0 \\
\bottomrule
\end{tabular}
\end{table}

H2 is established primarily by protocol: Table~\ref{tab:h2h3-results}(b) invokes the same checkpoint for the same canonical key in different task compositions, with zero additional Mani-Unit demonstrations and no task-trajectory update. The success-rate difference $\Delta$ measures context sensitivity; similar rates support stability but are not required for reuse because different prefixes may produce different physical states.

We replace circular ``selection accuracy'' with selection regret, $\max_k\widehat q_{z_i k}-\widehat q_{z_i k_i^*}$, and separately count invalid Mani-Unit keys or compatibility decisions.

\subsubsection{Agent Robustness}

Finally, we replace the frozen multimodal agent while holding the Mani-Unit vocabulary, executor checkpoints, capability registry, prompts, retry budget, and test configurations fixed. At least one proprietary and one open-source multimodal model are evaluated on Task-A. We compare the generated Mani-Unit keys and criteria with manually defined references, and compare visual verification with independently annotated ground truth.

\begin{table}[t]
\centering
\small
\setlength{\tabcolsep}{3pt}
\caption{Robustness to the multimodal agent. All entries are placeholders.}
\label{tab:agent-robustness}
\begin{tabular}{lcccc}
\toprule
Agent & Mani-Unit F1 & Key acc. & Ver. acc. & Task SR \\
\midrule
Proprietary & -- & -- & -- & -- \\
Open-source & -- & -- & -- & -- \\
\bottomrule
\end{tabular}
\end{table}

\bibliography{aaai2027}

\begin{thebibliography}{25}
\providecommand{\natexlab}[1]{#1}

\bibitem[{Ahn et~al.(2022)Ahn, Brohan, Brown, Chebotar et~al.}]{ahn2022saycan}
Ahn, M.; Brohan, A.; Brown, N.; Chebotar, Y.; et~al. 2022.
\newblock Do As {I} Can, Not As {I} Say: Grounding Language in Robotic
  Affordances.
\newblock arXiv:2204.01691.

\bibitem[{Brohan et~al.(2023)Brohan, Brown, Carbajal, Chebotar, Chen,
  Choromanski, Ding, Driess, Dubey, Finn et~al.}]{brohan2023rt2}
Brohan, A.; Brown, N.; Carbajal, J.; Chebotar, Y.; Chen, X.; Choromanski, K.;
  Ding, T.; Driess, D.; Dubey, A.; Finn, C.; et~al. 2023.
\newblock {RT-2}: Vision-Language-Action Models Transfer Web Knowledge to
  Robotic Control.
\newblock arXiv:2307.15818.

\bibitem[{Chen et~al.(2025)Chen, Chen, Chen, Cai, Liu, Li, Liang, Lin, Ge, Gu
  et~al.}]{chen2025robotwin}
Chen, T.; Chen, Z.; Chen, B.; Cai, Z.; Liu, Y.; Li, Z.; Liang, Q.; Lin, X.; Ge,
  Y.; Gu, Z.; et~al. 2025.
\newblock {RoboTwin} 2.0: A Scalable Data Generator and Benchmark with Strong
  Domain Randomization for Robust Bimanual Robotic Manipulation.
\newblock arXiv:2506.18088.

\bibitem[{Dietterich(2000)}]{dietterich2000hierarchical}
Dietterich, T.~G. 2000.
\newblock Hierarchical Reinforcement Learning with the {MAXQ} Value Function
  Decomposition.
\newblock \emph{Journal of Artificial Intelligence Research}, 13: 227--303.

\bibitem[{Driess et~al.(2023)Driess, Xia, Sajjadi et~al.}]{driess2023palme}
Driess, D.; Xia, F.; Sajjadi, M. S.~M.; et~al. 2023.
\newblock {PaLM-E}: An Embodied Multimodal Language Model.
\newblock In \emph{Proceedings of the 40th International Conference on Machine
  Learning}, volume 202 of \emph{Proceedings of Machine Learning Research},
  8469--8488.

\bibitem[{Hu et~al.(2026)Hu, Shridhar, Lu, Shah, Chiang, Tan, and
  Xie}]{hu2026matters}
Hu, J.; Shridhar, M.; Lu, C.; Shah, D.; Chiang, H.-T.~L.; Tan, J.; and Xie, A.
  2026.
\newblock What Matters in Orchestrating Robot Policies: A Systematic Study of
  Hierarchical {VLA} Agents.
\newblock arXiv:2606.10267.

\bibitem[{Huang et~al.(2022)Huang, Xia, Xiao, Chan, Liang, Florence, Zeng,
  Tompson, Mordatch, Chebotar, Sermanet, Brown, Jackson, Luu, Levine, Hausman,
  and Ichter}]{huang2022innermonologue}
Huang, W.; Xia, F.; Xiao, T.; Chan, H.; Liang, J.; Florence, P.; Zeng, A.;
  Tompson, J.; Mordatch, I.; Chebotar, Y.; Sermanet, P.; Brown, N.; Jackson,
  T.; Luu, L.; Levine, S.; Hausman, K.; and Ichter, B. 2022.
\newblock Inner Monologue: Embodied Reasoning through Planning with Language
  Models.
\newblock In \emph{Proceedings of the Conference on Robot Learning}, volume
  205, 1769--1782.

\bibitem[{James et~al.(2020)James, Ma, Arrojo, and Davison}]{james2020rlbench}
James, S.; Ma, Z.; Arrojo, D.~R.; and Davison, A.~J. 2020.
\newblock {RLBench}: The Robot Learning Benchmark and Learning Environment.
\newblock \emph{IEEE Robotics and Automation Letters}, 5(2): 3019--3026.

\bibitem[{Jiang et~al.(2023)Jiang, Gupta, Zhang, Wang, Dou, Chen, Fei-Fei,
  Anandkumar, Zhu, and Fan}]{jiang2023vima}
Jiang, Y.; Gupta, A.; Zhang, Z.; Wang, G.; Dou, Y.; Chen, Y.; Fei-Fei, L.;
  Anandkumar, A.; Zhu, Y.; and Fan, L. 2023.
\newblock {VIMA}: Robot Manipulation with Multimodal Prompts.
\newblock In \emph{Proceedings of the 40th International Conference on Machine
  Learning}, volume 202 of \emph{Proceedings of Machine Learning Research},
  14975--15022.

\bibitem[{Kaelbling, Littman, and Cassandra(1998)}]{kaelbling1998planning}
Kaelbling, L.~P.; Littman, M.~L.; and Cassandra, A.~R. 1998.
\newblock Planning and Acting in Partially Observable Stochastic Domains.
\newblock \emph{Artificial Intelligence}, 101(1--2): 99--134.

\bibitem[{Khazatsky et~al.(2024)Khazatsky, Pertsch, Nair, Balakrishna
  et~al.}]{khazatsky2024droid}
Khazatsky, A.; Pertsch, K.; Nair, S.; Balakrishna, A.; et~al. 2024.
\newblock {DROID}: A Large-Scale In-The-Wild Robot Manipulation Dataset.
\newblock In \emph{Proceedings of Robotics: Science and Systems}.

\bibitem[{Kim et~al.(2024)Kim, Pertsch, Karamcheti, Xiao, Balakrishna, Nair,
  Rafailov, Foster, Lam, Sanketi et~al.}]{kim2024openvla}
Kim, M.~J.; Pertsch, K.; Karamcheti, S.; Xiao, T.; Balakrishna, A.; Nair, S.;
  Rafailov, R.; Foster, E.; Lam, G.; Sanketi, P.; et~al. 2024.
\newblock {OpenVLA}: An Open-Source Vision-Language-Action Model.
\newblock arXiv:2406.09246.

\bibitem[{Li et~al.(2024)Li, Liang, Wang, Luo, Chen, Liao, Wei, Deng, Xu,
  Zhang, Wang, Liu, Fu, Bao, Chen, Shi, Yang, and Guo}]{li2024cogact}
Li, Q.; Liang, Y.; Wang, Z.; Luo, L.; Chen, X.; Liao, M.; Wei, F.; Deng, Y.;
  Xu, S.; Zhang, Y.; Wang, X.; Liu, B.; Fu, J.; Bao, J.; Chen, D.; Shi, Y.;
  Yang, J.; and Guo, B. 2024.
\newblock {CogACT}: A Foundational Vision-Language-Action Model for Synergizing
  Cognition and Action in Robotic Manipulation.
\newblock arXiv:2411.19650.

\bibitem[{Liang et~al.(2023)Liang, Huang, Xia, Xu, Hausman, Ichter, Florence,
  and Zeng}]{liang2023code}
Liang, J.; Huang, W.; Xia, F.; Xu, P.; Hausman, K.; Ichter, B.; Florence, P.;
  and Zeng, A. 2023.
\newblock Code as Policies: Language Model Programs for Embodied Control.
\newblock In \emph{Proceedings of the IEEE International Conference on Robotics
  and Automation}, 9493--9500.

\bibitem[{Liu et~al.(2026{\natexlab{a}})Liu, Chen, An, Liu, Zhang, Gu, Li, Guo,
  Chen, Liu, Hou, Zhao, Zhou, Heng, and Zhang}]{liu2025hybridvla}
Liu, J.; Chen, H.; An, P.; Liu, Z.; Zhang, R.; Gu, C.; Li, X.; Guo, Z.; Chen,
  S.; Liu, M.; Hou, C.; Zhao, M.; Zhou, K.; Heng, P.-A.; and Zhang, S.
  2026{\natexlab{a}}.
\newblock {HybridVLA}: Collaborative Diffusion and Autoregression in a Unified
  Vision-Language-Action Model.
\newblock In \emph{International Conference on Learning Representations}.

\bibitem[{Liu et~al.(2026{\natexlab{b}})Liu, Liu, Chen, Guo, Hou, Gu, Yu, Mi,
  Zhang, Che, Tang, Heng, and Zhang}]{last0}
Liu, Z.; Liu, J.; Chen, H.; Guo, Z.; Hou, C.; Gu, C.; Yu, J.; Mi, X.; Zhang,
  R.; Che, Z.; Tang, J.; Heng, P.-A.; and Zhang, S. 2026{\natexlab{b}}.
\newblock {LaST0}: Latent Spatio-Temporal Chain-of-Thought for Robotic
  Vision-Language-Action Model.
\newblock arXiv:2601.05248.

\bibitem[{Luo et~al.(2026)Luo, Chen, Wu, Sui, Liu, Gu, Liu, Feng, Yu, Gu, Jia,
  Heng, and Zhang}]{deepvisionvla}
Luo, Y.; Chen, H.; Wu, Z.; Sui, B.; Liu, J.; Gu, C.; Liu, Z.; Feng, Q.; Yu, J.;
  Gu, S.; Jia, P.; Heng, P.-A.; and Zhang, S. 2026.
\newblock Look Before Acting: Enhancing Vision Foundation Representations for
  Vision-Language-Action Models.
\newblock arXiv:2603.15618.

\bibitem[{Mees et~al.(2022)Mees, Hermann, Rosete-Beas, and
  Burgard}]{mees2022calvin}
Mees, O.; Hermann, L.; Rosete-Beas, E.; and Burgard, W. 2022.
\newblock {CALVIN}: A Benchmark for Language-Conditioned Policy Learning for
  Long-Horizon Robot Manipulation Tasks.
\newblock \emph{IEEE Robotics and Automation Letters}, 7(3): 7327--7334.

\bibitem[{{Open X-Embodiment Collaboration}(2024)}]{openx2024}
{Open X-Embodiment Collaboration}. 2024.
\newblock Open {X-Embodiment}: Robotic Learning Datasets and {RT-X} Models.
\newblock In \emph{Proceedings of the IEEE International Conference on Robotics
  and Automation}.

\bibitem[{{Physical Intelligence} et~al.(2025){Physical Intelligence}, Black,
  Brown, Darpinian, Dhabalia, Driess, Esmail, Equi, Finn, Fusai
  et~al.}]{physicalintelligence2025pi05}
{Physical Intelligence}; Black, K.; Brown, N.; Darpinian, J.; Dhabalia, K.;
  Driess, D.; Esmail, A.; Equi, M.; Finn, C.; Fusai, N.; et~al. 2025.
\newblock {$\pi_{0.5}$}: A Vision-Language-Action Model with Open-World
  Generalization.
\newblock arXiv:2504.16054.

\bibitem[{Qu et~al.(2025)Qu, Song, Chen, Yao, Ye, Ding, Wang, Gu, Zhao, Wang,
  and Li}]{qu2025spatialvla}
Qu, D.; Song, H.; Chen, Q.; Yao, Y.; Ye, X.; Ding, Y.; Wang, Z.; Gu, J.; Zhao,
  B.; Wang, D.; and Li, X. 2025.
\newblock {SpatialVLA}: Exploring Spatial Representations for
  Visual-Language-Action Model.
\newblock arXiv:2501.15830.

\bibitem[{Sutton, Precup, and Singh(1999)}]{sutton1999between}
Sutton, R.~S.; Precup, D.; and Singh, S. 1999.
\newblock Between {MDPs} and Semi-{MDPs}: A Framework for Temporal Abstraction
  in Reinforcement Learning.
\newblock \emph{Artificial Intelligence}, 112(1--2): 181--211.

\bibitem[{Zhang and Shah(2026)}]{zhang2026visualverification}
Zhang, M.; and Shah, D. 2026.
\newblock Visual Verification Enables Inference-Time Steering and Autonomous
  Policy Improvement.
\newblock In \emph{Proceedings of Robotics: Science and Systems}.

\bibitem[{Zhao et~al.(2025)Zhao, Lu, Kim, Fu, Zhang, Wu, Li, Ma, Han, Finn,
  Handa, Liu, Xiang, Wetzstein, and Lin}]{zhao2025cotvla}
Zhao, Q.; Lu, Y.; Kim, M.~J.; Fu, Z.; Zhang, Z.; Wu, Y.; Li, Z.; Ma, Q.; Han,
  S.; Finn, C.; Handa, A.; Liu, M.-Y.; Xiang, D.; Wetzstein, G.; and Lin, T.-Y.
  2025.
\newblock {CoT-VLA}: Visual Chain-of-Thought Reasoning for
  Vision-Language-Action Models.
\newblock In \emph{Proceedings of the IEEE/CVF Conference on Computer Vision
  and Pattern Recognition}.

\bibitem[{Zheng et~al.(2026)Zheng, Li, Wang, Liu, Kang, Feng, Zheng, Zou, Chen,
  Zeng et~al.}]{zheng2026xvla}
Zheng, J.; Li, J.; Wang, Z.; Liu, D.; Kang, X.; Feng, Y.; Zheng, Y.; Zou, J.;
  Chen, Y.; Zeng, J.; et~al. 2026.
\newblock {X-VLA}: Soft-Prompted Transformer as a Scalable Cross-Embodiment
  Vision-Language-Action Model.
\newblock In \emph{International Conference on Learning Representations}.

\end{thebibliography}
\end{document}